\documentclass{article}

\usepackage{PRIMEarxiv}

\usepackage[utf8]{inputenc} 
\usepackage[T1]{fontenc}    
\usepackage{hyperref}       
\usepackage{url}            
\usepackage{booktabs}       
\usepackage{amsfonts}       
\usepackage{nicefrac}       
\usepackage{microtype}      
\usepackage{lipsum}
\usepackage{fancyhdr}       
\usepackage{graphicx}       

\usepackage{tabularx}
\usepackage{amsmath}
\usepackage[ruled,vlined]{algorithm2e}

\title{Bayesian Active Summarization}

\author{
  Alexios Gidiotis \\
  School of Informatics \\
  Aristotle University of Thessaloniki \\
  Thessaloniki, Greece \\
  \texttt{gidiotis@csd.auth.gr} \\
   \And
  Grigorios Tsoumakas \\
  School of Informatics \\
  Aristotle University of Thessaloniki \\
  Thessaloniki, Greece \\
  \texttt{greg@csd.auth.gr} \\
}

\begin{document}
\maketitle

\begin{abstract}
Bayesian Active Learning has had significant impact to various NLP problems, but nevertheless it's application to text summarization has been explored very little. We introduce Bayesian Active Summarization (BAS), as a method of combining active learning methods with state-of-the-art summarization models. Our findings suggest that BAS achieves better and more robust performance, compared to random selection, particularly for small and very small data annotation budgets. Using BAS we showcase it is possible to leverage large summarization models to effectively solve real-world problems with very limited annotated data.
\end{abstract}

\keywords{Active learning \and abstractive text summarization \and Bayes methods \and Monte Carlo methods \and natural language processing \and deep learning}

\section{Introduction}
\label{sec:introdction}
Modern Natural Language Processing (NLP) methods based on deep neural networks have achieved remarkable performance in several different tasks \cite{Vaswani2017AttentionNeed, Radford2019LanguageLearners, Devlin2018Bert:Understanding, liu2019roberta, Lewis2019BART:Comprehension,  Raffel2020ExploringTransformer}. Such performance levels are usually achieved by scaling up deep neural networks to millions or even billions of parameters. Scaling, in turn, requires extremely high computational capacity and large training datasets.

Abstractive text summarization is an NLP task that has drawn extensive attention in recent deep learning work, with some very significant achievements \cite{Lewis2019BART:Comprehension, Zhang2020PEGASUS:Summarization, zaheer2020big}. State-of-the-art summarization models generally depend on large supervised datasets to achieve good performance and generalize to unseen data. Summarization datasets are usually document collections, each accompanied by some form of summary, typically written by humans.

Although numerous such datasets are available in the literature \cite{Napoles2012AnnotatedGigaword, Hermann2015TeachingComprehend, Narayan2018DontSummarization, Grusky2018Newsroom:Strategies}, this is not the case in many practical applications, since constructing large supervised datasets is very expensive and time consuming. Collecting good quality training data in large amounts and annotating them for summarization would be costly for many small businesses trying to adopt summarization technology to solve problems in their respective domains. This cost can be particularly high if {\em domain expertise} is required for annotation, which is true for many use cases such as financial and legal documents.

Active learning methods have been widely adopted in an effort to reduce deep learning data requirements for various tasks \cite{Houlsby2011BayesianLearning, Gal2017DeepData}. Strategically selecting for annotation the most informative samples has proven to be more effective than random selection when the budget for annotating data is small \cite{Siddhant2020DeepStudy}. Active learning has also been applied to NLP problems but it has rarely been explored from a summarization perspective\cite{Zhang2009ActiveSummarization}.

We present Bayesian Active Summarization (BAS), an approach for applying active learning to abstractive text summarization aiming to mitigate the data dependence of summarization models. BAS iteratively alternates between annotating and training, in an effort to maximize the gains from a limited data annotation budget. Based on previous work \cite{Gidiotis2021Uncertainty-AwareSummarization}, we apply Bayesian deep learning methods with Monte Carlo (MC) dropout \cite{Gal2016DropoutLearning} to quantify summarization uncertainty, and use it to select data samples for annotation.

We empirically show that BAS is more data efficient than random selection, and achieves better overall performance when the annotation budget is small. More specifically, we conducted experiments using the PEGASUS summarization model \cite{Zhang2020PEGASUS:Summarization} and managed to reach 95\% of the performance of a PEGASUS model trained on the full XSum dataset\cite{Narayan2018DontSummarization} using less than 150 training samples.

Finally, we analyze BAS with regard to computational cost, in an effort to identify the effects different design choices have. This analysis gives us insights into the trade-off between performance and computational complexity, helping us understand how to scale BAS effectively.

The rest of this paper is structured as follows. Section \ref{sec:related_work} is a review of relevant work in active learning, Bayesian methods and their application to NLP. In Section \ref{sec:bayesian_uncertainty} we briefly introduce Bayesian uncertainty and it's extensions to summarization models. Then, in Section \ref{sec:active_sum} we present in detail the main BAS algorithm and in Section \ref{sec:practical_considerations} some practical considerations concerning scalability and robustness. In Section \ref{sec:exp_setup} we describe our experimental setup and in Section \ref{sec:results} we discuss our main findings, including an analysis of BAS from different angles. We conclude with some final remarks in Section \ref{sec:conclusion}.

\section{Related work}
\label{sec:related_work}
A number of works have applied active learning methods to NLP problems. \cite{Shen2018DeepRecognition} use active learning for named entity recognition (NER), by selecting and annotating the samples with the lowest predicted probability. Most notably, \cite{Siddhant2020DeepStudy} empirically study multiple different active learning methods, focusing on sentence classification and NER. Various works combine BERT \cite{Devlin2018Bert:Understanding} with different active learning methods on NLP tasks like text classification \cite{Ein-Dor2020ActiveStudy}, NER \cite{Liu2020LTP:Recognition} and natural language understanding (NLU) \cite{Griehaber2020Fine-tuningLearning}.

Bayesian active learning methods have been successfully applied to various problems, ranging from computer vision \cite{Kendall2017WhatVision, Litjens2017AAnalysis, Gal2017DeepData} to NLU \cite{Griehaber2020Fine-tuningLearning} and NER \cite{Siddhant2020DeepStudy}. Most of the aforementioned methods use BALD \cite{Houlsby2011BayesianLearning} variations and MC dropout \cite{Gal2016DropoutLearning} in order to acquire samples for annotation.

Closely related to our work, \cite{Lyu2020YouAnswering} uses the BLEUVar disagreement \cite{Xiao2020WatTransformers} between samples generated with MC dropout as an uncertainty metric for selecting samples to annotate. They mainly focus on the question answering task, and show that performance can be improved by training on a subset of the most uncertain samples.

In stark contrast, active learning methods have seen limited adoption in summarization. \cite{Zhang2009ActiveSummarization} proposes an active learning method that utilizes additional resources, and selects documents based on their similarity with PowerPoint slides from corresponding presentations. They then annotate the selected documents and use them to train an extractive summarization model.

Another interesting work by \cite{Schick2020Few-ShotTraining} tries to reduce summarization data requirements without the use of active learning. They propose augmenting the inputs to the PEGASUS model \cite{Zhang2020PEGASUS:Summarization} in order to make fine-tuning more effective. They achieve considerable improvements in few-shot summarization performance on multiple benchmark datasets. Nevertheless, their work assumes that only a few, between 10 and 100, annotated samples are available while we focus on actively selecting samples for annotation. Consequently, our methods are not directly comparable to their approach. 

Finally, our previous work \cite{Gidiotis2021Uncertainty-AwareSummarization} suggests using MC dropout to estimate summarization models’ uncertainty. Similar to \cite{Lyu2020YouAnswering}, multiple stochastic summaries are generated and the BLEUVar disagreement is computed between those summaries, as an uncertainty estimate. Inspired by that work, we apply summarization uncertainty in the context of active learning, to acquire highly uncertain documents at each annotation round.

\section{Bayesian uncertainty}
\label{sec:bayesian_uncertainty}
In this section we briefly introduce some key concepts about Bayesian uncertainty, which is the foundation of our document selection strategy.

\subsection{Monte Carlo dropout}
Standard deep learning models are notorious for their highly confident predictions, even for inputs lying further away from their learned distribution \cite{Gal2016DropoutLearning, Xiao2020WatTransformers}. In contrast, Bayesian probabilistic models can explicitly model uncertainty based on the models’ predictive distribution variance. Finding the model's predictive distribution involves deriving the entire {\em posterior} distribution over all model parameters $\theta$ given training data $X$ and $Y$  (Equation \ref{eq:posterior}), and at test time making predictions by integrating over all possible $\theta$ values (Equation \ref{eq:integration}).

\begin{equation}
\label{eq:posterior}
P(\theta | X,Y) = \frac{P(Y | X, \theta)P(\theta)}{P(Y | X)}
\end{equation}

\begin{equation}
\label{eq:integration}
P(\hat{y} | x,X,Y) = \int P(\hat{y} | x, \theta) P(\theta | X,Y) d\theta
\end{equation}

Although integrating over all possible $\theta$ values is intractable for deep neural networks, alternative methods can be used to approximate it during inference. Monte Carlo dropout \cite{Gal2016DropoutLearning} is one such method that performs multiple stochastic forward passes with dropout \cite{Srivastava2014Dropout:Overfitting} turned on at test time, which is equivalent to drawing from the model's predictive distribution. It is a practical method to approximate Bayesian inference, readily applicable to various deep learning models.

\subsection{Application to summarization}
When applied to text summarization \cite{Gidiotis2021Uncertainty-AwareSummarization}, MC dropout can be used to estimate uncertainty. The uncertainty estimation process for summarization models involves two steps. First, $N$ stochastic summaries are sampled for a given input $x$ by running forward passes with different dropout masks. Then, the disagreement between summaries, measured by BLEU Variance (BLEUVar) \cite{Xiao2020WatTransformers} between all summary pairs, can be used to estimate summarization uncertainty. In practice, BLEUVar is computed by summing the squared complement of BLEU \cite{Papineni2002BLEU:Translation} among all summary pairs generated with MC dropout as shown in Equation \ref{eq:bleuvar}.

\begin{equation}
\small
\label{eq:bleuvar}
BLEUVar = \frac{1}{N(N-1)}\sum_{i=1}^{N} \sum_{j \neq i}^{N} (1 - BLEU(y_i,y_j))^2
\end{equation}

\section{Active summarization}
\label{sec:active_sum}
The main objective of Bayesian active summarization (BAS) is to train a summarization model that achieves competitive performance, but requires less supervised data for training. Since creating large numbers of samples for summarization training can be particularly difficult and costly, we focus on training budgets of only a few hundreds of annotated samples. 

Active learning methods, and particularly Bayesian ones, are known to have significant computational overheads. When applied to the abstractive summarization task, which also has high resource requirements, such methods can lead to very high computational costs. A secondary objective of this work is to develop a practical and effective method that is also resource efficient.

For BAS we follow the well established active learning paradigm, that strategically selects
data to annotate over alternating rounds of labeling and training \cite{Cohn1996ActiveModels, Houlsby2011BayesianLearning, Siddhant2020DeepStudy}. The full active summarization algorithm is shown in Algorithm \ref{alg:as_algorithm}. In Table \ref{tab:notation} we collect the notation used in this section.

We start off with a learner $M_{init}$, which is a standard abstractive summarization model. Initially, we have a pool of unlabeled documents $U$ and an empty set of labeled documents $L$. Since our goal is to achieve strong performance with as few annotated documents as possible, it makes sense for the initial learner to be some pre-trained model, such as BART \cite{Lewis2019BART:Comprehension} or PEGASUS \cite{Zhang2020PEGASUS:Summarization}, but in principle any neural summarization model could work.

During training we also need a separate annotated set, which will be used for validation and early stopping. Following what was suggested in \cite{Siddhant2020DeepStudy}, we keep a separate validation set of $v$ labeled documents, which is proportional to our total annotation budget. It would be unrealistic to have a few hundred examples for training but thousands of annotated examples for validation.

First, we warm-start our learner by training on a small sample of $s_0$ randomly selected and annotated (summarized) documents from $U$. This part is important because it essentially ``introduces" the summarization task to the learner. In practice, a pre-trained Transformer based model would require a few dozens of examples for this initial learning stage. The initial training documents are also added to the labeled set $L$ and removed from $U$.

The rest of the learning iteratively alternates between labeling (summarizing) and training, where each full iteration is called a learning step. The process is repeated for multiple steps, until we reach satisfactory performance or we exhaust our annotation budget.

The labeling phase in each learning step is split into two parts. First, we have the learner generate $n$ stochastic summaries with MC dropout for each document in the unlabeled pool $U$, and based on these summaries we compute the BLEUVar uncertainty score for each document. Then, the top $s\ll |U|$ documents with the highest uncertainty are selected and target summaries are retrieved for them.

After the labeling phase, we proceed to the training phase. The documents annotated in the previous phase are added to the labeled set $L$, and the learner is trained on the whole labeled set. At each learning step, we train the learner from scratch in order to avoid overfitting the data samples collected in earlier steps \cite{Hu2019ActiveFeedback}. We also make sure to train the learner for a sufficient number of epochs, with early stopping based on performance on a validation set $V$ consisting of $v$ samples, in order to avoid underfitted learners \cite{Mukhoti2018OnLearning}.


\begin{algorithm}[htb]
\SetAlgoLined
 $L \leftarrow \emptyset$\;
 $i \leftarrow 0$\;
 $L \leftarrow sample(U, s_0)$\;
 $V \leftarrow sample(U, v)$\;
 $M_{0} \leftarrow train(M_{init}, L, V)$\;
 \While{$size(L) \leq b$}{
  $B \leftarrow genSums(M_i, U, n)$\;
  $S \leftarrow rankSums(B, s)$\;
  $L \leftarrow L \cup S$\;
  $U \leftarrow U \setminus S$\;
  $i\texttt{++}$\;
  $M_{i} \leftarrow train(M_{init}, L, V)$\;
 }
 \KwResult{$M_i$}
\caption{Bayesian Active Summarization}
\label{alg:as_algorithm}
\end{algorithm}

\begin{table}
\caption{Notation table for BAS}
\label{tab:notation}
\begin{center}
\begin{tabular}{cc}
\\\hline
{\textbf{symbol}} & {\textbf{desciption}} \\
\hline
$L$ & labeled set\\
$l$ & number of samples in the labeled set\\
$U$ & unlabeled set\\
$u$ & number of samples in the unlabeled set\\
$V$ & validation set\\
$v$ & number of validation samples\\
$K$ & sampled set to be ranked by uncertainty \\
$k$ & number of samples to be ranked\\
$S$ & set of documents selected based on high uncertainty \\
$s$ & number of documents selected in each step\\
$n$ & number of MC dropout samples \\
$b$ & training budget \\
$s_0$ & number of documents for ``warm-start" training \\
$M_{init}$ & pre-trained summarization model \\
$M_{0}$ & summarization model after ``warm-start" training\\
$M_{i}$ & summarization model trained at step $i$\\
\hline
\end{tabular}
\end{center}
\end{table}

\section{Practical considerations}
\label{sec:practical_considerations}
In this section we discuss practical considerations that arise when trying to apply active summarization in real world scenarios. These considerations concern important decisions we need to make with each applications' specific requirements in mind, and have a great impact on BAS's practical performance, cost, scalability and robustness.

\subsection{Trading off precision for speed}
\label{subsec:k_sampling}
Active learning methods are usually trading off data collection and annotation costs for increased computational costs. Although collecting and annotating data is usually the most costly and time consuming part, having very high computational costs is generally sub-optimal.

Let's assume that we have a training budget of b samples and the cost of generating a single summary is constant and equal to $C_{sum}$ while the cost of computing BLEU for a pair of texts is equal to $C_{bl}$. Equation \ref{eq:BLEUVar_cost} gives us the cost of ranking and selecting the $s$ most uncertain samples out of the full unlabeled set of $u$ documents.

\begin{equation}
\label{eq:BLEUVar_cost}
C(u,n) = un[C_{sum} + (n-1)C_{bl}]
\end{equation}

If we also assume that the cost of each training step is constant and equal to $C_{train}$ and the cost of the initial, warm-start, training is $C_{train0}$, the total cost of BAS is given by Equation \ref{eq:bas_cost}.

\begin{equation}
\label{eq:bas_cost}
C_{BAS}(u,n,s) = \frac{b}{s}(C(u,n) + C_{train}) + C_{train0}
\end{equation}

We can see that the BAS cost is a function of the number of documents to rank $u$, the MC dropout samples $n$ and the number of selected samples $s$. Since $n$ and $s$ are generally expected to be much smaller than $u$, which is the size of the entire unlabeled set, and the total budget $b$ is predefined, then the total BAS cost is mostly dependent on $u$.

In many practical applications we expect the unlabeled set $U$ to be rather large. This is a common pattern in all active learning methods, but in the case of BAS we need to take into account that modern abstractive summarization models have a high computational cost and we need to perform multiple forward passes for each input sample. With these considerations in mind we can see that the computational complexity of Algorithm \ref{alg:as_algorithm} can become prohibitively high.

To address high computational complexity, we relax our assumption and instead of trying to find the documents with the highest uncertainty from the entire unlabeled pool we randomly sample a set $K$ consisting of $k$ documents from the unlabeled pool $U$, where $k\ll|U|$. Then in the labeling phase, we use MC dropout to generate summaries and estimate uncertainty only for the documents in $K$. Once we select the top $s\ll k$ documents from $K$, the remaining documents that were not selected are returned to the unlabeled pool and can be selected again at a later learning step. After this modification, the cost of BAS is no longer dependent on $U$ as can be seen in Equation \ref{eq:bas_cost_sample}. The adjusted BAS algorithm is shown in Algorithm \ref{alg:as_algorithm_with_k}.

\begin{align}
\label{eq:bas_cost_sample}
\begin{split}
C(k,n) = kn[C_{sum} + (n-1)C_{bl}] \\
C_{BAS}(k,n,s) = \frac{b}{s}(C(k,n) + C_{train}) + C_{train0}
\end{split}
\end{align}

\begin{algorithm}[htb]
\SetAlgoLined
 $L \leftarrow \emptyset$\;
 $i \leftarrow 0$\;
 $L \leftarrow sample(U, s_0)$\;
 $V \leftarrow sample(U, v)$\;
 $M_{0} \leftarrow train(M_{init}, L, V)$\;
 \While{$size(L) \leq b$}{
  $K \leftarrow sample(U, k)$\;
  $B \leftarrow genSums(M_i, K, n)$\;
  $S \leftarrow rankSums(B, s)$\;
  $L \leftarrow L \cup S$\;
  $U \leftarrow U \setminus S$\;
  $i\texttt{++}$\;
  $M_{i} \leftarrow train(M_{init}, L, V)$\;
 }
 \KwResult{$M_i$}
\caption{Bayesian Active Summarization (adjusted)}
\label{alg:as_algorithm_with_k}
\end{algorithm}

To further illustrate the importance of the sampling step, we refer the reader to Table \ref{tab:k_run_times}, which shows an experimental study over different sample sizes. We show the average computational time which can decomposed into scoring time ($C(k,n)$) and training time ($C_{train}$). It is clear that increasing the sample size $k$ leads to exponential increase for the scoring time, taking up a significant proportion of the total time and out-weights the training time. One can imagine that in the extreme case where $k=u$, the computational time for each step will be very high even for medium size datasets.

With the introduction of this sampling step, we effectively stop worrying about finding the most uncertain samples in the pool, and instead we just look for samples of sufficiently high uncertainty. By limiting the number of documents participating in the labeling phase, we effectively reduce each steps' computational complexity significantly. The parameter $k$, which is the number of documents we sample at each step, controls the trade-off between precision and speed. High $k$ values guarantee the selected documents will have very high uncertainty, but also mean we will have to process a large number of documents during each steps' labeling phase. On the other hand, low $k$ values mean only a few documents will be processed at each step, but would lead to the selected documents having lower uncertainty on average.

One additional benefit that comes with randomly selecting $k$ samples to rank, is that it allows us to sidestep a common active learning issue. Selecting the highest uncertainty samples from the entire dataset can easily result in acquiring a batch of very similar examples at each learning step. This problem is known to hurt performance and generalization for active learning methods \cite{Kirsch2019BatchBALD:Learning}.

In Section \ref{sec:scaling_discussion} we dive deeper into the impact of this trade-off between efficiency and performance, and we empirically explore the effects different $k$ values have.

\subsection{Dealing with noisy data}
\label{sec:outliers}
Another practical implication that comes with selecting very high uncertainty samples is that we run the risk of them being noisy or of poor quality. As discussed in \cite{Gidiotis2021Uncertainty-AwareSummarization}, particularly in noisy datasets, documents with extremely high uncertainty could be problematic, and summarization performance could be improved if we eliminate them from the training set. Examples of problematic documents include ones that are malformed, written in a different language, or completely missing meaningful content.

Although the sampling strategy discussed in Section \ref{sec:practical_considerations} helps with the problem of noisy samples, it does not address it completely. Such samples could be thought of as outliers in the uncertainty distribution and as such they should be removed from the training data. Especially when the data budget is low, keeping noisy samples could be detrimental to the overall performance, because they could end up accounting for a significant proportion of the training dataset.

We introduce a heuristic method that uses a simple uncertainty threshold to remove documents of disproportionately high uncertainty. This heuristic enables us to filter out most noisy documents and ensure the selected documents are of relatively good quality.

\section{Experimental setup}
\label{sec:exp_setup}
In order to assess BAS's effectiveness, an experimental study was conducted. Here, we describe our experimental setup including the data and model used for the study as well as the learning details. We aim to simulate a real-world scenario with a low data annotation budget and as a consequence all experimental decisions are made under that assumption.

\subsection{Data}
XSum \cite{Narayan2018DontSummarization} is a dataset of 227k news articles from BBC covering a wide variety of topics. Each article is paired with a human written, single-sentence summary. We used the XSum version openly available in the Hugging Face datasets repository\footnote{\href{https://huggingface.co/datasets}{https://huggingface.co/datasets}}.

Since our emphasis is on small data budgets, we set the total data annotation budget to 900 samples and use the whole 200k article training set as the unlabeled pool. Each time a document is selected we retrieve its target summary and add it to the labeled set. Out of the total data budget, $b=u=800$ samples will be used for training  and the rest $v=100$ samples will be left to the validation set, to be used for early stopping evaluation.

At the end of each learning step the generated models' performance is evaluated on the full XSum test set consisting of 11k articles. This evaluation step is not part of the active summarization algorithm described in Section \ref{sec:active_sum}, and diverges from the low resource setup. We purposely made this decision in order to facilitate comparisons with existing methods on the XSum dataset and to verify the overall effectiveness of BAS.

\subsection{Model}
PEGASUS \cite{Zhang2020PEGASUS:Summarization} is a Transformer based sequence-to-sequence summarization model achieving very strong performance on multiple well established summarization benchmarks. The models' encoder and decoder are built of 16 Transformer blocks each. PEGASUS is pre-trained on the C4 and HugeNews datasets, on a sentence infilling task. We have used the open-sourced and pre-trained PEGASUS version found in the Hugging Face models repository\footnote{\href{https://huggingface.co/models}{https://huggingface.co/models}}.

In order to perform Bayesian inference with PEGASUS, we follow the same approach used in \cite{Gidiotis2021Uncertainty-AwareSummarization} by enabling dropout during inference for all Transformer blocks.

\subsection{Learning details}
Here we will present the learning details and parameters used when applying BAS in our experiments. We start off with the openly available, pre-trained PEGASUS model and we initially train it on $s_0=50$ randomly sampled and annotated documents. Then we proceed to run multiple learning steps as described in Section \ref{sec:active_sum} until we exhaust our 800 samples training budget. At the end of each round, we evaluate the performance of the resulting model on the XSum test set. Each experiment was repeated three times with different random seeds in an effort to alleviate the effects of randomness, and all metrics were averaged across all three runs.

In the labeling phase, we use MC dropout with $n=10$ \cite{Gidiotis2021Uncertainty-AwareSummarization} to compute the uncertainty of the model trained on the previous learning step. When generating summaries with the model, we use standard beam search with a beam size of 3, which is a good trade-off between performance and speed. In our main experiments we used $s=10$ with different values for $k$. Also, when selecting samples for annotation, we ignore all documents with uncertainty higher than 0.96, as explained in Section \ref{sec:outliers}. This heuristic threshold was determined after experimentation. 

In the training phase, we start each time from the pre-trained PEGASUS model and train on the whole labeled set $L$ retrieved so far. In each step, the model is trained for 10 epochs at most, using early stopping with a patience of 4 epochs. We avoid extensively fine-tuning hyper-parameters because in a real world scenario, with limited data, extensive fine-tuning would result in severe overfitting. For the majority of the training hyper-parameters the values from the original PEGASUS paper \cite{Zhang2020PEGASUS:Summarization} are used except for the batch size, which was set to 6 in order to keep resource requirements low. With this hyper-parameter setup we managed to run the entire learning on a single Nvidia T4 GPU.

\section{Results}
\label{sec:results}
The experimental results presented in this section are organized in the following way. First, we go through the process of tuning BAS, in an attempt to achieve a good balance between effectiveness and computational complexity. Then, we evaluate the performance of BAS and compare it with a baseline that follows the standard supervised learning paradigm of randomly selecting and annotating samples.

\subsection{Tuning for performance and scalability}
\label{sec:scaling_discussion}
As described in Section \ref{subsec:k_sampling}, our method samples $k$ documents instead of calculating BLEUVar over the entire unlabeled set, which would be prohibitively expensive in many cases. Sampling allows us to greatly reduce BAS's computational complexity, since scoring lots of documents with MC dropout is a very costly operation.

The number of sampled documents $k$ is a hyper-parameter that can significantly affect the performance and scalability of our method. We assess the effects of different values by running experiments with $k$ set to $50$, $100$, $200$ and $500$. We then compare the outcomes of these experiments in terms of summarization performance and computational cost.

\begin{figure*}
    \makebox[\textwidth][c]{\includegraphics[width=1.2\textwidth]{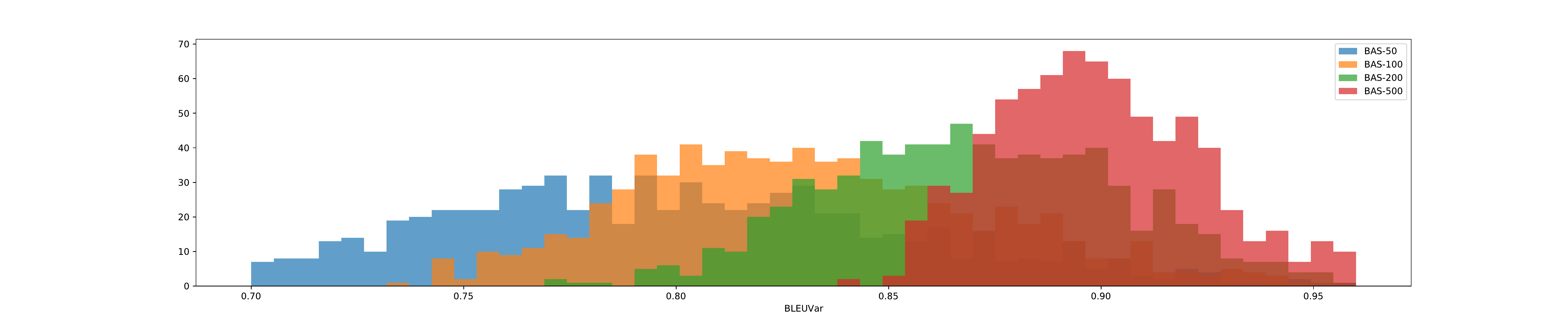}}
    \caption{Histogram of BLEUVar uncertainty for the documents selected for annotation with different $k$ values.}
    \label{fig:bleu_hist}
\end{figure*}

In general, smaller values of $k$ result in faster run times as shown in Table \ref{tab:k_run_times}, but the samples selected at each step have lower uncertainty on average. Figure \ref{fig:bleu_hist} illustrates the BLEUVar uncertainty scores distributions for different $k$ values. We observe that for higher $k$ values there is a significant shift in the distribution of the selected documents towards higher uncertainty values.

\begin{table}
\caption{\label{tab:k_run_times} Run times (in seconds) for different $k$ values. We show total run time per learning step, uncertainty scoring time per step for BAS approaches and total run times.}
\begin{center}
\begin{tabular}{ccccc}
\\\hline
 {\textbf{k}} & {\textbf{Total/step}} & {\textbf{Scoring/step}} & {\textbf{Total}} & {\textbf{Avg. BLEUVar}}\\
\hline
50 & 1,144 & 150 & 87,000 & 0.79\\
100 & 1,196 & 210 & 90,900 & 0.84\\
200 & 1,434 & 370 & 108,775 & 0.87\\
500 & 2,113 & 1,280 & 159,675 & 0.90\\
\hline
\end{tabular}
\end{center}
\end{table}

\begin{figure*}
    \makebox[\textwidth][c]{\includegraphics[width=1.3\textwidth]{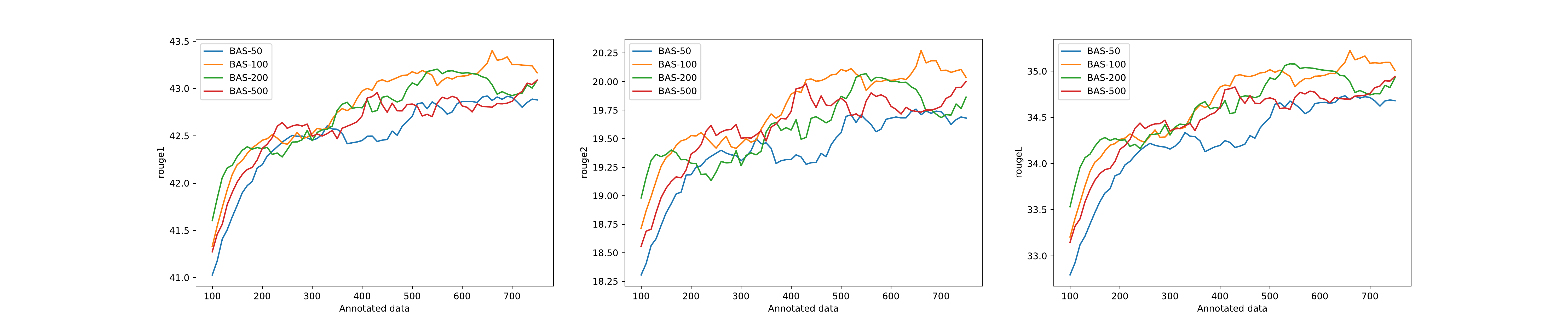}}
    \caption{Performance curves of BAS with different values for $k$ vs random acquisition. We plot ROUGE-1, ROUGE-2 and ROUGE-L F-scores as a function of the total annotated data for a total of 75 learning steps.}
    \label{fig:bas_k}
\end{figure*}

In Figure \ref{fig:bas_k} we show the performance curves for ROUGE-1, ROUGE-2 and ROUGE-L F-score \cite{Lin2004Rouge:Summaries}, obtained with different $k$ values. Our first observation is that all curves converge to similar performance at the later learning stages but exhibit different behaviors at the early stages. We can see that increasing $k$ yields better performance early on, when the data budget is smaller, but the gains are not very significant towards the end.

The improvement in the performance curves can be attributed to the higher average uncertainty of the documents selected using higher $k$ values. As the average uncertainty of the selected samples gets lower, BAS selects samples closer to the full datasets' mean giving up the advantages of active learning.

Based on our findings in this Section, we remark that using $k$ values smaller than 100 leads to a drop in performance and thus should be avoided . Also, increasing $k$ higher than 200 does not bring a convincing improvement in performance but increases computational costs considerably. We conclude that scoring $100$-$200$ samples and selecting the ones with the highest uncertainty offers a good trade-off between performance and efficiency. For the rest of our experiments, we will be using $k=100$ and $k=200$.

\subsection{Performance}
\label{sec:performance_results}
\begin{figure*}
    \makebox[\textwidth][c]{\includegraphics[width=1.3\textwidth]{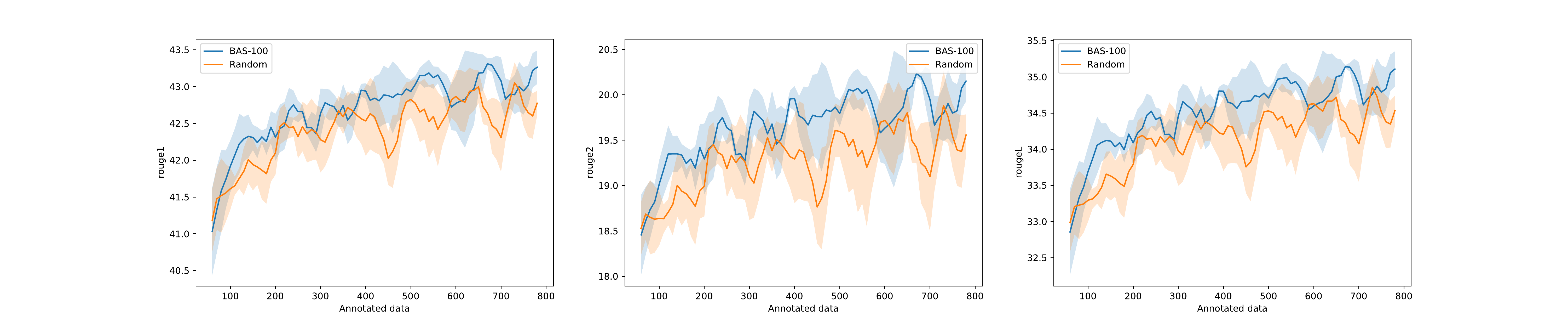}}
    \caption{Performance curves of BAS-100 vs random acquisition for a total of 75 learning steps. We plot the ROUGE-1, ROUGE-2 and ROUGE-L F-scores, averaged across three runs, as a function of the total annotated data.}
    \label{fig:xsum_k100}
\end{figure*}

\begin{figure*}
    \makebox[\textwidth][c]{\includegraphics[width=1.3\textwidth]{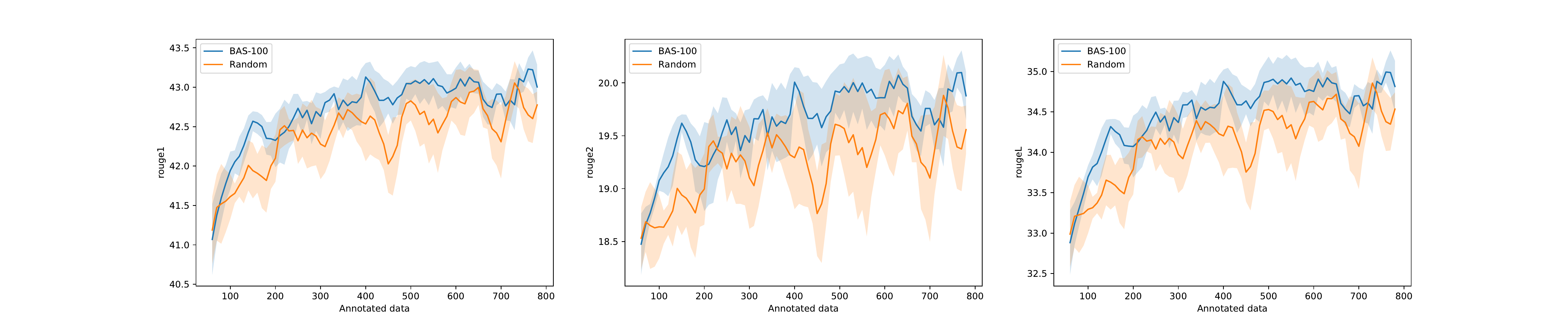}}
    \caption{Performance curves of BAS-200 vs random acquisition for a total of 75 learning steps. We plot the ROUGE-1, ROUGE-2 and ROUGE-L F-scores, averaged across three runs, as a function of the total annotated data.}
    \label{fig:xsum_k200}
\end{figure*}

We plot the ROUGE-1, ROUGE-2 and ROUGE-L F-score performance for each learning step in Figures \ref{fig:xsum_k100} ($k=100$) and \ref{fig:xsum_k200} ($k=200$), and compare the curves of Bayesian Active Summarization (BAS) against a baseline that uses random  selection  at  each  step. All curves are averaged across three runs. 

In Table \ref{tab:best_metrics} we show the best performance achieved by the different approaches in terms of ROUGE. For the sake of comparison, we also show the performance achieved by PEGASUS when trained on the entire training set (PEGASUS full) with standard supervised learning\footnote{re-evaluated for consistency} as well as the performance without any task specific training (PEGASUS pre-trained).

\begin{table*}
\caption{\label{tab:best_metrics} ROUGE-1, ROUGE-2 and ROUGE-L best scores achieved by different approaches for small (800 samples) and very small (150 samples) training budgets. All scores are computed on the entire validation set.}
\begin{center}
\setlength\tabcolsep{0.2cm}
\begin{tabular}{ccccccc}
\\\hline
& \multicolumn{3}{c}{\textbf{Small \small(800 samples)}} & \multicolumn{3}{c}{\textbf{Very small \small(150 samples)}}\\
\hline
& {\textbf{ROUGE-1}} & {\textbf{ROUGE-2}} & {\textbf{ROUGE-L}} & {\textbf{ROUGE-1}} & {\textbf{ROUGE-2}} & {\textbf{ROUGE-L}}\\
\hline
Random & 43.25 $\pm0.32$ & 20.07 $\pm0.36$ & 35.02 $\pm0.34$ & 42.06 $\pm0.34$ & 19.14 $\pm0.30$ & 33.77 $\pm0.33$\\
BAS-100 & \textbf{43.40 $\pm0.28$} & \textbf{20.32 $\pm0.29$} & \textbf{35.26 $\pm0.29$} & 42.39 $\pm0.42$ & 19.45 $\pm0.29$ & 34.20 $\pm0.40$\\
BAS-200 & 43.38 $\pm0.24$ & 20.24 $\pm0.25$ & 35.11 $\pm0.24$ & \textbf{42.55 $\pm0.28$} & \textbf{19.59 $\pm0.20$} & \textbf{34.31 $\pm0.25$}\\
\hline
PEGASUS pre-trained & 17.84 & 2.65 & 12.71 & - & - & -\\
PEGASUS full & 44.90 & 23.33 & 37.74 & - & - & -\\
\hline
\end{tabular}
\end{center}
\end{table*}

Overall, BAS performs better than random selection as can be seen by the performance curves. More specifically, the performance curves of BAS-100 and BAS-200 are higher compared to random selection across all metrics. As the number of annotated samples increases we can see both BAS and random start converging to similar performance although BAS is still slightly higher than random selection. Also, in terms of ROUGE, BAS's best performance is higher than random, with BAS-100 having a slight edge over BAS-200 on that aspect.

The biggest advantage of BAS is when the annotated dataset is very small, for example less than 150 samples, where it clearly outperforms random by a considerable margin. Also, for very small data budgets BAS-200 is better than BAS-100. Based on our experiments, with BAS-200 it is possible to get close to 95\% of the performance of a PEGASUS model trained on the full XSum with less than 150 annotated samples. In comparison to BAS, we can see that with the same training budget, random selection achieves 93\% of that performance, while only 39\% of it is covered by the pre-trained model. Finally, using 5 times more data we can reach almost 97\%, which is only a slight improvement if we take into account the significant increase in data annotation cost.

Being able to achieve good performance with such a small data budget is an extremely useful property in many real-world applications, since  collecting and annotating even a few hundreds of training samples can be a challenge. Our findings in this experiment suggest that with BAS large summarization models such as PEGASUS could be applied effectively to solve few-shot problems.

Although the percentage difference between BAS and random selection is not particularly large, this is mainly due to the fact that PEGASUS is already a very effective pre-trained model. We argue that since this is a strong model, it is much harder to improve its performance by a large margin. Nevertheless, BAS is still a useful tool that allows us to get more value out of a small training budget. 


In Figures \ref{fig:xsum_k100} and \ref{fig:xsum_k200} we also plot the standard deviation over the three runs for each individual curve. We observe BAS exhibits smaller standard deviation across multiple runs compared to random selection, leading us to the conclusion that BAS would be more robust and less affected by stochastic factors. This observation is more clearly visible in Table \ref{tab:best_metrics} where we show the standard deviations of the different approaches for all metrics. BAS-200 has significantly lower standard deviation compared to BAS-100, which suggests that increasing $k$  can lead to more robust solutions.

Robustness is particularly important in practical active learning setups, where getting extremely unlucky when selecting data samples to be annotated could lead to inferior performance. In practice, repeating the data selection process with a different random seed is not really an option due to the fixed budget, so it is crucial to have a method more likely to find a good solution.

\section{Conclusion}
\label{sec:conclusion}
In this work, we explored active learning in the context of abstractive text summarization. Although active learning methods have had significant impact on various NLP problems, applications to text summarization have been very limited. We introduced BAS, as a way of combining active learning methods with state-of-the-art summarization models. BAS is, to the best of our knowledge, the first attempt to apply Bayesian active learning to abstractive text summarization.

Our main findings suggest that indeed BAS can be an effective way for applying active learning to summarization, outperforming naive random selection in several ways. BAS achieves stronger performance and has better learning curves on small and very small annotation budgets compared to random selection. In addition, it leads to more robust learning as it reduces the risk of being extremely unlucky and selecting bad samples for annotation. At the same time, it allows for identifying and eliminating noisy samples that could hurt performance.

The main advantage of BAS, and active learning methods in general, is the ability to achieve strong performance with very few annotated data samples. As shown in our experiments, we managed to reach 95\% of the performance of the fully trained PEGASUS model, using less than 150 training samples. This finding can have a significant impact in many real-world applications where collecting large datasets is very costly.

In addition, we performed an experimental analysis of different BAS setups in an attempt to better understand the effect of different design decisions with regards to performance and computational efficiency. We found that selecting the most uncertain documents from a small sub-sample of the full dataset yields satisfactory performance and scales well. 


Although our findings suggest Bayesian Active Learning is a promising approach for improved abstractive summarization, we are barely scratching the surface of this interesting and very little explored topic. We hope our work will spark a broader discussion and will be a starting point for further exploring active learning methods on the task of text summarization.

\bibliographystyle{unsrt}  
\bibliography{references}

\begin{thebibliography}{10}

\bibitem{Vaswani2017AttentionNeed}
Ashish Vaswani, Noam Shazeer, Niki Parmar, Jakob Uszkoreit, Llion Jones,
  Aidan~N. Gomez, Lukasz Kaiser, and Illia Polosukhin.
\newblock {Attention is all you need}.
\newblock In {\em Advances in Neural Information Processing Systems}, pages
  5998--6008, 2017.

\bibitem{Radford2019LanguageLearners}
Alec Radford, Jeffrey Wu, Rewon Child, David Luan, Dario Amodei, and Ilya
  Sutskever.
\newblock {Language Models are Unsupervised Multitask Learners}.
\newblock {\em OpenAI Blog}, 1(8), 2019.

\bibitem{Devlin2018Bert:Understanding}
Jacob Devlin, Ming-Wei Chang, Kenton Lee, and Kristina Toutanova.
\newblock {Bert: Pre-training of deep bidirectional transformers for language
  understanding}.
\newblock {\em arXiv preprint arXiv:1810.04805}, 2018.

\bibitem{liu2019roberta}
Yinhan Liu, Myle Ott, Naman Goyal, Jingfei Du, Mandar Joshi, Danqi Chen, Omer
  Levy, Mike Lewis, Luke Zettlemoyer, and Veselin Stoyanov.
\newblock {Roberta: A robustly optimized bert pretraining approach}.
\newblock {\em arXiv preprint arXiv:1907.11692}, 2019.

\bibitem{Lewis2019BART:Comprehension}
Mike Lewis, Yinhan Liu, Naman Goyal, Marjan Ghazvininejad, Abdelrahman Mohamed,
  Omer Levy, Ves Stoyanov, and Luke Zettlemoyer.
\newblock {BART: Denoising sequence-to-sequence pre-training for natural
  language generation, translation, and comprehension}.
\newblock In {\em Proceedings of the 58th Annual Meeting of the Association for
  Computational Linguistics}, pages 7871--7880. Association for Computational
  Linguistics, 2019.

\bibitem{Raffel2020ExploringTransformer}
Colin Raffel, Noam Shazeer, Adam Roberts, Katherine Lee, Sharan Narang, Michael
  Matena, Yanqi Zhou, Wei Li, and Peter~J. Liu.
\newblock {Exploring the limits of transfer learning with a unified
  text-to-text transformer}.
\newblock {\em Journal of Machine Learning Research}, 21, 2020.

\bibitem{Zhang2020PEGASUS:Summarization}
Jingqing Zhang, Yao Zhao, Mohammad Saleh, and Peter~J Liu.
\newblock {PEGASUS: Pre-training with Extracted Gap-sentences for Abstractive
  Summarization}.
\newblock In {\em 37th International Conference on Machine Learning, ICML
  2020}, pages 11328--11339. PMLR, 2020.

\bibitem{zaheer2020big}
Manzil Zaheer, Guru Guruganesh, Avinava Dubey, Joshua Ainslie, Chris Alberti,
  Santiago Ontanon, Philip Pham, Anirudh Ravula, Qifan Wang, and Li~Yang.
\newblock {Big bird: Transformers for longer sequences}.
\newblock {\em arXiv preprint arXiv:2007.14062}, 2020.

\bibitem{Napoles2012AnnotatedGigaword}
Courtney Napoles, Matthew Gormley, and Benjamin Van~Durme.
\newblock {Annotated gigaword}.
\newblock In {\em Proceedings of the Joint Workshop on Automatic Knowledge Base
  Construction and Web-scale Knowledge Extraction}, pages 95--100, 2012.

\bibitem{Hermann2015TeachingComprehend}
Karl~Moritz Hermann, Tomas Kocisky, Edward Grefenstette, Lasse Espeholt, Will
  Kay, Mustafa Suleyman, and Phil Blunsom.
\newblock {Teaching machines to read and comprehend}.
\newblock In {\em Advances in Neural Information Processing Systems}, pages
  1693--1701. Curran Associates, Inc., 2015.

\bibitem{Narayan2018DontSummarization}
Shashi Narayan, Shay~B. Cohen, and Mirella Lapata.
\newblock {Don't give me the details, just the summary! Topic-aware
  convolutional neural networks for extreme summarization}.
\newblock In {\em Proceedings of the 2018 Conference on Empirical Methods in
  Natural Language Processing, EMNLP 2018}, pages 1797--1807, Brussels,
  Belgium, 2018. Association for Computational Linguistics.

\bibitem{Grusky2018Newsroom:Strategies}
Max Grusky, Mor Naaman, and Yoav Artzi.
\newblock {Newsroom: A Dataset of 1.3 Million Summaries with Diverse Extractive
  Strategies}.
\newblock In {\em Proceedings of the 2018 Conference of the North American
  Chapter of the Association for Computational Linguistics: Human Language
  Technologies}, pages 708--719, 2018.

\bibitem{Houlsby2011BayesianLearning}
Neil Houlsby, Ferenc Huszar, Zoubin Ghahramani, and Mate Lengyel.
\newblock {Bayesian active learning for classification and preference
  learning}, 2011.

\bibitem{Gal2017DeepData}
Yarin Gal, Riashat Islam, and Zoubin Ghahramani.
\newblock {Deep Bayesian active learning with image data}.
\newblock In {\em 34th International Conference on Machine Learning, ICML
  2017}, volume~3, pages 1183--1192. PMLR, 2017.

\bibitem{Siddhant2020DeepStudy}
Aditya Siddhant and Zachary~C. Lipton.
\newblock {Deep Bayesian active learning for natural language processing:
  Results of a large-scale empirical study}.
\newblock In {\em Proceedings of the 2018 Conference on Empirical Methods in
  Natural Language Processing, EMNLP 2018}, pages 2904--2909, Brussels,
  Belgium, 2020. Association for Computational Linguistics.

\bibitem{Zhang2009ActiveSummarization}
Justin~Jian Zhang and Pascale Fung.
\newblock {Active learning of extractive reference summaries for lecture speech
  summarization}.
\newblock In {\em Proceedings of the 2nd Workshop on Building and Using
  Comparable Corpora from Parallel to Non-parallel Corpora}, pages
  23--undefined. Association for Computational Linguistics, 2009.

\bibitem{Gidiotis2021Uncertainty-AwareSummarization}
Alexios Gidiotis and Grigorios Tsoumakas.
\newblock {Uncertainty-Aware Abstractive Summarization}, 2021.

\bibitem{Gal2016DropoutLearning}
Yarin Gal and Zoubin Ghahramani.
\newblock {Dropout as a Bayesian approximation: Representing model uncertainty
  in deep learning}.
\newblock In {\em 33rd International Conference on Machine Learning, ICML
  2016}, volume~3, pages 1050--1059. PMLR, 2016.

\bibitem{Shen2018DeepRecognition}
Yanyao Shen, Hyokun Yun, Zachary~C. Lipton, Yakov Kronrod, and Animashree
  Anandkumar.
\newblock {Deep active learning for named entity recognition}.
\newblock In {\em 6th International Conference on Learning Representations,
  ICLR 2018 - Conference Track Proceedings}, 2018.

\bibitem{Ein-Dor2020ActiveStudy}
Liat Ein-Dor, Alon Halfon, Ariel Gera, Eyal Shnarch, Lena Dankin, Leshem
  Choshen, Marina Danilevsky, Ranit Aharonov, Yoav Katz, and Noam Slonim.
\newblock {Active Learning for BERT: An Empirical Study}.
\newblock In {\em Proceedings of the 2020 Conference on Empirical Methods in
  Natural Language Processing, EMNLP 2020}, 2020.

\bibitem{Liu2020LTP:Recognition}
Mingyi Liu, Zhongjie Wang, Zhiying Tu, and Xiaofei Xu.
\newblock {LTP: a new active learning strategy for BERT-CRF based named entity
  recognition}, 2020.

\bibitem{Griehaber2020Fine-tuningLearning}
Daniel Grie{\ss}haber, Johannes Maucher, and Ngoc~Thang Vu.
\newblock {Fine-tuning BERT for Low-Resource Natural Language Understanding via
  Active Learning}.
\newblock In {\em Proceedings of the 28th International Conference on
  Computational Linguistics. 2020}, 2020.

\bibitem{Kendall2017WhatVision}
Alex Kendall and Yarin Gal.
\newblock {What uncertainties do we need in Bayesian deep learning for computer
  vision?}
\newblock In {\em Advances in Neural Information Processing Systems}, volume
  2017-December, pages 5580--5590. Curran Associates, Inc., 2017.

\bibitem{Litjens2017AAnalysis}
Geert Litjens, Thijs Kooi, Babak~Ehteshami Bejnordi, Arnaud Arindra~Adiyoso
  Setio, Francesco Ciompi, Mohsen Ghafoorian, Jeroen~A.W.M. van~der Laak, Bram
  van Ginneken, and Clara~I. S{\'{a}}nchez.
\newblock {A survey on deep learning in medical image analysis}, 2017.

\bibitem{Lyu2020YouAnswering}
Zhihao Lyu, Danier Duolikun, Bowei Dai, Yuan Yao, Pasquale Minervini, Tim~Z
  Xiao, and Yarin Gal.
\newblock {You Need Only Uncertain Answers: Data Efficient Multilingual
  Question Answering}.
\newblock In {\em Workshop on Uncertainty and Robustness in Deep Learning},
  2020.

\bibitem{Xiao2020WatTransformers}
Tim~Z. Xiao, Aidan~N. Gomez, and Yarin Gal.
\newblock {Wat zei je? Detecting out-of-distribution translations with
  variational transformers}, 2020.

\bibitem{Schick2020Few-ShotTraining}
Timo Schick and Hinrich Sch{\"{u}}tze.
\newblock {Few-Shot Text Generation with Pattern-Exploiting Training}.
\newblock {\em arXiv preprint arXiv:2012.11926}, 2020.

\bibitem{Srivastava2014Dropout:Overfitting}
Nitish Srivastava, Geoffrey Hinton, Alex Krizhevsky, Ilya Sutskever, and Ruslan
  Salakhutdinov.
\newblock {Dropout: A simple way to prevent neural networks from overfitting}.
\newblock {\em Journal of Machine Learning Research}, 15:1929--1958, 2014.

\bibitem{Papineni2002BLEU:Translation}
Kishore Papineni, Salim Roukos, Todd Ward, and Wei-Jing Zhu.
\newblock {BLEU: a method for automatic evaluation of machine translation}.
\newblock In {\em Proceedings of the 40th annual meeting on association for
  computational linguistics}, pages 311--318. Association for Computational
  Linguistics, 2002.

\bibitem{Cohn1996ActiveModels}
David~A. Cohn, Zoubin Ghahramani, and Michael~I. Jordan.
\newblock {Active learning with statistical models}.
\newblock {\em Journal of Artificial Intelligence Research}, 4, 1996.

\bibitem{Hu2019ActiveFeedback}
Peiyun Hu, Zachary~C. Lipton, Anima Anandkumar, and Deva Ramanan.
\newblock {Active learning with partial feedback}.
\newblock In {\em 7th International Conference on Learning Representations,
  ICLR 2019}, 2019.

\bibitem{Mukhoti2018OnLearning}
Jishnu Mukhoti, Pontus Stenetorp, and Yarin Gal.
\newblock {On the importance of strong baselines in bayesian deep learning},
  2018.

\bibitem{Kirsch2019BatchBALD:Learning}
Andreas Kirsch, Joost van Amersfoort, and Yarin Gal.
\newblock {BatchBALD: Efficient and diverse batch acquisition for deep Bayesian
  active learning}.
\newblock In {\em Advances in Neural Information Processing Systems},
  volume~32, 2019.

\bibitem{Lin2004Rouge:Summaries}
Chin-Yew Lin.
\newblock {Rouge: A package for automatic evaluation of summaries}.
\newblock In {\em Proceedings of the 2004 Workshop on Text Summarization
  Branches Out, Post Conference Workshop of ACL}, 2004.

\end{thebibliography}

\end{document}